\pdfoutput=1

\documentclass[11pt]{article}

\usepackage{EMNLP2022}

\usepackage{times}
\usepackage{latexsym}

\usepackage[T1]{fontenc}

\usepackage[utf8]{inputenc}

\usepackage{microtype}
\usepackage{multirow}
\usepackage{tabularx}
\usepackage{amsmath}
\usepackage{amssymb}
\usepackage{bm}
\usepackage{subfigure}
\usepackage{arydshln}
\usepackage[switch]{lineno}
\usepackage{helvet} 
\usepackage{courier}  
\usepackage{graphicx} 
\usepackage{natbib}  
\usepackage{caption} 

\usepackage{times}
\usepackage{latexsym}

\usepackage{inconsolata}

\title{Continual Learning of Neural Machine Translation \\ within Low Forgetting Risk Regions}

\author{Shuhao Gu\textsuperscript{\rm 1,2}\thanks{$^*$The work was done when the first author was an intern at Tencent.}, \textbf{Bojie Hu\textsuperscript{\rm 3}}, Yang Feng\textsuperscript{\rm 1,2}\thanks{Corresponding author: Yang Feng. \newline \indent ~~ Reproducible code: https://github.com/ictnlp/LFR-NMT. }\\ 
\textsuperscript{\rm 1} Key Laboratory of Intelligent Information Processing,\\ Institute of Computing Technology, Chinese Academy of Sciences (ICT/CAS)\\
\textsuperscript{\rm 2} University of Chinese Academy of Sciences\\
\textsuperscript{\rm 3} Tencent Minority-Mandarin Translation, Beijing, China\\
{ \{gushuhao19b,fengyang\}@ict.ac.cn, \indent bojiehu@tencent.com}}

\begin{document}
\maketitle
\begin{abstract}
This paper considers continual learning of large-scale pretrained neural machine translation model without accessing the previous training data or introducing model separation. 
We argue that the widely used regularization-based methods, which perform multi-objective learning with an auxiliary loss, suffer from the misestimate problem and cannot always achieve a good balance between the previous and new tasks.
To solve the problem, we propose a two-stage training method based on the local features of the real loss.
We first search low forgetting risk regions, where the model can retain the performance on the previous task as the parameters are updated, to avoid the catastrophic forgetting problem.
Then we can continually train the model within this region only with the new training data to fit the new task.
Specifically, we propose two methods to search the low forgetting risk regions, which are based on the curvature of loss and the impacts of the parameters on the model output, respectively.
We conduct experiments on domain adaptation and more challenging language adaptation tasks, and the experimental results show that our method can achieve significant improvements compared with several strong baselines.

\end{abstract}

\section{Introduction}
The current large-scale pretrained neural machine translation (NMT) models, such as GMNMT~\cite{JohnsonSLKWCTVW17}, mBART50-nn~\cite{abs-2008-00401}, and M2M-100~\cite{FanBSMEGBCWCGBL21}, are generally trained with large amounts of data from different domains and languages, so the model can learn good semantic representation and mapping relationship at the same time. 
Based on such models, we hope that they can continually acquire new knowledge from new translation tasks, e.g., new domains and languages, while preserving the previously learned knowledge, i.e., continual learning (CL).
However, catastrophic forgetting is the biggest challenge of continual learning, which means the model will forget the previously learned knowledge while learning new knowledge~\cite{gu-feng-2020-investigating}. 
One solution to avoid forgetting is to mix the previous and the new training data to train the model jointly. 
However, considering that the training data of the large-scale pre-trained NMT model is usually very large, this method will bring more training consumption and consume more energy. Besides, we sometimes cannot access the previous data due to data privacy or storage limitation.


\begin{figure}[t]
    \centering
    \includegraphics[width=\columnwidth]{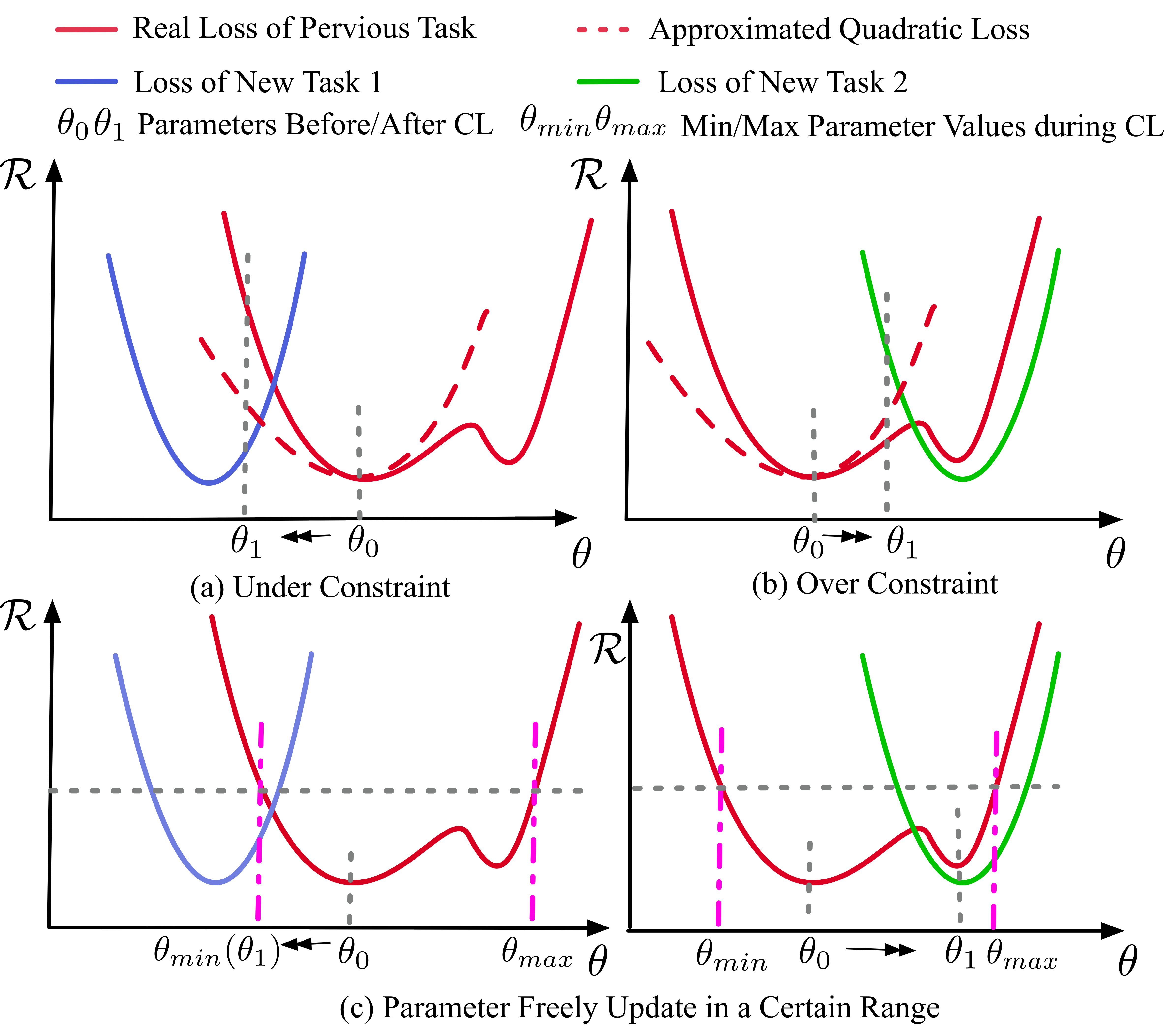}
    \caption{An illustration to indicate the difference between the regularization-based method and our method. $\mathcal{R}$ denotes the task loss. $\theta$ denotes the model parameters. $\rightarrow$ and $\leftarrow$ denote the parameter update direction during continual learning. $[\theta_{min}, \theta_{max}]$ denotes the low forgetting risk regions.}
    \label{fig:intro}
\end{figure}

To solve this problem, many researchers have made their attempts. 
Some work avoids catastrophic forgetting by constructing pseudo data and mixing them with the new task data for joint training~\cite{KimR16,KoERCGGFKD20,LiuWF21}. 
However, these methods must require that the previous and new tasks are similar, and can not be directly applied to tasks with large differences, such as learning a totally different new language. 
Some work introduces extra task-specific parameters and only updates these parameters with the new task data~\cite{BapnaF19,EscolanoCF21}. 
On the one hand, this will increase the size of the model. 
On the other hand, the task-specific parameters lead to model separation, which makes the model must know which task the input sentence belongs to, otherwise, the translation quality will degrade significantly~\cite{AharoniG20}.
In contrast, the regularization-based methods~\cite{KhayrallahTDK18,ThompsonGKDK19} don't have such limitations and are more flexible in practice.
These methods perform multi-objective learning with an extra penalty item on the parameters, which aims to approximate the real loss on the previous task, and is usually in quadratic form.
However, as illustrated in Figure~\ref{fig:intro} (a) and (b), the approximated loss (the red dashed line) is convex and symmetric, but the real loss (the red solid line) is not necessarily so in most cases, which may lead to the under or over constraint problems. 
Besides, the multi-objective learning can only make the parameters converge to a region, where the gradients produced by the two losses are equal in size and opposite in direction, i.e., gradients reach balance. However, this does not guarantee that the values of the two losses are small in this region, which means the model may still suffer from the forgetting problem.


Instead of performing multi-objective learning with the approximated loss, we propose a two-stage continual learning method based on the local features of the real loss around the initial parameters.
In the first stage, we aim to search the low forgetting risk regions as the parameter update regions, where the performance of the previous task will not degrade significantly when the parameters are updated. 
Therefore, we can constrain the parameters to such regions to avoid catastrophic forgetting.
In the second stage, the parameters will be freely updated in the searched regions only guided by the gradients of the new task, while the update outside these regions will be forbidden, as illustrated in Figure~\ref{fig:intro} (c). 
To achieve this, we can use some data related to the previous task to help us find such regions, e.g., the validation set data of the previous translation task, which is usually small-scale and easy to obtain. 
We propose two methods to search the low forgetting risk regions. 
The first method is to find out which parameters have the least impact on the previous task loss. 
We use the curvature of the loss curve as an indicator and the parameter with small curvature indicates that the loss curve is relatively flat, so the parameters can be updated more greedily. 
The second method is based on the impacts of parameters on the model output of the previous task.
We define an objective function, which tries to maximize the size of update regions as much as possible while restricting the change of model output on the previous task, to help the model learn the update regions automatically.
We conduct experiments on the domain adaptation and more challenging language adaptation tasks, and the results show that our method can achieve significant improvements compared with several strong baselines.

Our contribution can be summarized as follows:
\begin{itemize}
    \item We propose two methods to search the low forgetting risk regions of neural machine translation without accessing the previous training data.
    \item We conduct experiments on the domain adaptation task and the more challenging language adaptation task of continual learning, and our method can bring significant improvements. 
    \item We prove that our method can also achieve good performance when combined with part of the original training data. 
    
\end{itemize}

\section{Background}
In this section, we report the background knowledge used in this paper: the Transformer~\cite{VaswaniSPUJGKP17} model, multilingual NMT, Hessian matrix, and Fisher information matrix.
\subsection{Transformer}
We denote the input sequence of symbols as $\mathbf{x}=(x_1,\ldots,x_{I})$ and the ground-truth sequence as $\mathbf{y}=(y_1,\ldots,y_{J})$.
The transformer model is based on the encoder-decoder architecture. The encoder is composed of $\mathnormal{N}$ identical layers. Each layer has two sublayers. The first is a multi-head self-attention sublayer, and the second is a fully connected feed-forward network. Both of the sublayers are followed by a residual connection operation and a layer normalization operation. 
The input sequence $\mathbf{x}$ will be first converted to a sequence of vectors $\mathbf{E}_x=[E_x[x_1];\ldots;E_x[x_J]]$ where $E_x[x_j]$ is the sum of word embedding and position embedding of the source word $x_j$. 
Then, this sequence of vectors will be fed into the encoder, and the output of the $\mathnormal{N}$-th layer will be taken as source state sequences. We denote it as $\mathbf{H}_\mathbf{x}$. 
The decoder is also composed of $\mathnormal{N}$ identical layers. In addition to the same kind of two sublayers in each encoder layer, the cross-attention sublayer is inserted between them, which performs multi-head attention over the output of the encoder. 
The final output of the $\mathnormal{N}$-th layer gives the target hidden states $\mathbf{S}=[\mathbf{s}_1;\ldots;\mathbf{s}_{K*}]$, where $\mathbf{s}_k$ is the hidden states of $y_k$. 
We can get the predicted probability of the $j$-th target word conditioned by the source sentence and the $j-1$ previous target words. The model is optimized by minimizing a cross-entropy loss of the ground-truth sequence with teacher forcing:
\begin{equation}\label{eq::loss}
    \mathcal{L}_{CE} = -\frac{1}{J} \sum_{j=1}^{J} \log p(y_j | \mathbf{y}_{<j}, \mathbf{x}; \mathbf{\theta}),
\end{equation}
where $J$ is the length of the target sentence and $\theta$ denotes the model parameters. 

\subsection{Multilingual Translation}
The multilingual neural machine translation (MNMT) model can translate between multiple different languages with a single model~\cite{JohnsonSLKWCTVW17}. 
Following~\citet{LiuGGLEGLZ20}, we add a particular language id token at the beginning of the source and target sentences, respectively, to indicate the language. 

\subsection{Hessian and Fisher Information Matrices}\label{sec.matrix}
\textbf{The Hessian Matrix} For a twice-differentiable loss $\mathcal{L}$, the Hessian matrix is the matrix of second-order derivatives of the loss function with respect to the weights, mathematically expressed as $\mathbf{H}=\nabla^{2}_{\theta} \mathcal{L}$. Intuitively, its role is to express the curvature of the loss around a given point $\theta$. The smaller the value, the 
"flatter" the loss is around the given point $\theta$, and vice versa. The flatter the region around $\theta$, the smaller the loss change when the value of $\theta$ is changed. 

\noindent \textbf{The Fisher Information Matrix}
The Fisher information matrix $\mathbf{F}$ of the model’s conditional distribution $P_{\mathbf{y}|\mathbf{x}}$ is defined as:
\begin{equation}
    \mathbf{F} = \mathrm{E}_{P_{\mathbf{x,y}}}[\nabla_{\theta}\log p_{\theta}(\mathbf{x,y})\nabla_\theta \log p_{\theta}(\mathbf{x,y})^{\top}]
\end{equation}
Intuitively, the role of the Fisher matrix is very similar to that of the Hessian Matrix.
If $\theta$ is an accurate set of parameters for the model, we can approximate the Hessian matrix at $\theta$ with the Fisher information matrix~\cite{ly2017tutorial}. 

In practice settings, we can approximate the Fisher information matrix by replacing the model distribution $P_{\mathbf{x,y}}$ with the empirical training distribution $\hat{Q}_{\mathbf{x,y}}$:
\begin{equation}\label{eq:fisher}
\begin{split}
    \hat{\mathbf{F}}& =  \mathrm{E}_{\hat{Q}_{\mathbf{x,y}}}[\nabla_{\theta}\log p_{\theta}(\mathbf{x,y})\nabla_\theta \log p_{\theta}(\mathbf{x,y})^{\top}] \\
    &= \frac{1}{N}\sum_{n=1}^{N}\nabla \log p_{\theta}(\mathbf{y}|\mathbf{x})\nabla \log p_{\theta}(\mathbf{y}|\mathbf{x})^{\top}
\end{split}
\end{equation}

\section{Method}
The goal of our method is to preserve the previously learned knowledge while adapting to the new tasks efficiently.
To achieve this, we propose a two-stage training method.
In the first stage, we try to depict the local features of the large-scale pre-trained NMT model around its initial parameters $\theta_0$.
Specifically, we want to find a region around $\theta_0$ with low forgetting risk, so we can retain the performance of the model on the previous task as the parameters are updated.
We can constrain the parameters within this region so that the model will not suffer from severe forgetting problem. 
In the second stage, the parameters are updated completely guided by the gradients produced by the new training data within this region, which can mitigate the under-fitting problem on the new task.  

\subsection{Search Parameter Update Region}
We propose two methods to search the low forgetting risk (LFR) regions around the initial parameters $\theta_0$. 
For the first method, we hope to find out which parameters to update have the least impact on the model loss of the previous task. 
To achieve this, we propose to use curvature as an indicator to help find the parameter update regions.
For the second method, we determine the LFR regions based on the impact of parameters on the model output of the previous task.
We propose an objective function to help the model learn the regions.

\textbf{Curvature-based Method} Intuitively, the curvature of the loss function measures how fast the loss changes as the parameter changes around the initial parameters $\theta_0$. 
Therefore, the parameters with small curvature can be safely updated without causing forgetting. 
As described in Section~\ref{sec.matrix}, the Hessian matrix is used to represent the curvature of the model loss, but it is almost impossible to obtain the exact Hessian matrix in practice. 
Therefore, we use the Fisher information matrix to approximate the Hessian matrix with Equation~\ref{eq:fisher}. 
Noting that we cannot access the previous training data, we use a small-scale validation set related to the previous task to compute the Fisher information matrix, which will be described in the experimental part. 

After getting the Fisher information matrix, we fix the top $\rho\%$ parameters of each module with large values, and then set the update regions for the rest of the parameters as $[\theta_{\min}, \theta_{\max}]$, which are defined as:
\begin{equation}
    \begin{split}
        & \theta_{\min} = \theta_0 - \lambda * |\theta_0|; \\
        & \theta_{\max} = \theta_0 + \lambda * |\theta_0|,
    \end{split}
\end{equation}
where $\theta_0$ denotes the parameter values of the pre-trained NMT model, and $\lambda$ is a hyper-parameter to control the size of the update region. 

\textbf{Output-based Method} 
In this method, we hope that the model can automatically learn the update region of parameters based on the impact of parameters on the model output of the previous task.
Intuitively, preserving the previously learned knowledge requires that the model output on the previous task should not change significantly after parameter update. 
Meanwhile, we also hope that the update regions should be as large as possible because the large regions will give the model more capacity to learn new tasks.
Following the above intuition, we can define the learning objective:
\begin{equation}
\begin{split}
    \mathcal{L}(\theta)=\frac{1}{N}\sum_{n=1}^{N}KL&(p(\mathbf{y}|\mathbf{x}, \theta)||p(\mathbf{y}|\mathbf{x}, \theta_0)) \\
   & - \frac{\alpha}{M} \sum_{i=1}^{M} (\theta_i-\theta_{0,i})^2,
\end{split}
\end{equation}
where $N$ denotes the amount of the training example, $KL$ denotes the KL-divergence, $\alpha$ is a hyper-parameter to control the ratio of the two terms, and $M$ denotes the total amount of the model parameters. 
The first term of the above objective function let the model output on the previous task stay as close as possible to the pre-trained model, which will discourage the parameter to change. 
While the second term encourages the model parameters to change more greedily, and maximize the size of regions as much as possible. 
These two items can be regarded as performing adversarial learning during learning the parameter update regions. 
Similar to the curvature-based method, we use a small-scale validation set related to the previous task instead of the data of the new task as the training data. 

After this learning process, we can get the updated model parameters $\theta_1$, then we define the update region $[\theta_{\min}, \theta_{\max}]$ as:
\begin{equation}
\begin{split}
    & \theta_{\min} = \min(\theta_0, \theta_1); \\
    & \theta_{\max} = \max(\theta_0, \theta_1).
\end{split}
\end{equation}

\subsection{Hard-Constrained Training}
After finding the parameter update regions, we continually train the model parameters within this region to learn the new translation tasks, i.e., to find:
\begin{equation}
\begin{split}
    \theta^{*} = & \arg \  \underset{\theta}{\min} \sum_{\mathbf{(x,y)\in \mathcal{D}_{N}}} \mathcal{L}_{CE}(\mathbf{y|x},\theta), \\
    & s.t. \ \theta_{\min} \leq \theta \leq \theta_{\max},
\end{split}
\end{equation}
where $\mathcal{D}_{N}$ denotes the training data of the new task. 
One may suspect that updating parameters only in a constrained region may lead to the insufficient ability of the model to fit the new translation tasks. 
However, considering the over-parameterization of the large-scale NMT model and the fact that many parameters have not learned sufficient knowledge~\cite{HoeflerABDP21}, the hard-constrained training can give model the capability to adapt to new tasks in most cases, even though the update range of parameters is limited.
The experimental results will also prove this.

\section{Experiments}
In this work, we perform continual learning on two tasks: the domain adaptation task and the more challenging language adaptation task. 
In the domain adaptation task, the language pairs of the new training data are already supported by the pre-trained NMT model, and the goal is to enable the model to support the translation of new domains. 
In the language adaptation task, the goal is to enable the model to support the translation of new languages, which are not seen during pretraining. 
Under the above two scenarios, we hope to retain the translation ability of the pre-trained MNMT model on the original translation task.

\subsection{The Pre-trained NMT Model}
In the experiments, we use the mBART50-nn~\cite{abs-2008-00401} model as our pre-trained NMT model, which is available in the fairseq library. The mBart50-nn is a many-to-many multilingual NMT model which can support the translation between 50 different languages. 
The encoder and decoder of mBART50-nn have 12 layers, respectively. The dimensionality of the model is set as 1024, and the attention module has 16 attention heads.
The dimensionality of the embedding layer and hidden states is set as 1024, while the inner-layer of the feed-forward network has dimensionality as 4096. The attention module has 16 attention heads both in the encoder and decoder. Besides, the model has a shared source-target vocabulary of about 250k tokens, and the model uses learned positional embeddings with the max token length set as 1024. 

\begin{table}[]
\centering
\begin{tabular}{lc|ccc}
\hline
\multicolumn{2}{c|}{Task} & Train & Valid & Test \\ \hline \hline
\multicolumn{1}{l|}{\begin{tabular}[c]{@{}l@{}}Multilingual\\ Translation\end{tabular}} & xx$\leftrightarrow$En & / & 997 & 1012 \\ \hline
\multicolumn{1}{l|}{\multirow{5}{*}{\begin{tabular}[c]{@{}l@{}}Domain\\ Adaptation\\(De$\rightarrow$En)\end{tabular}}} & IT & 0.22M & \multirow{5}{*}{2000} & \multirow{5}{*}{2000} \\
\multicolumn{1}{l|}{} & Law & 0.47M &  &  \\
\multicolumn{1}{l|}{} & Medical & 0.25M &  &  \\
\multicolumn{1}{l|}{} & Subtitles & 0.5M &  &  \\
\multicolumn{1}{l|}{} & Koran & 18K &  &  \\ \hline
\multicolumn{1}{l|}{\multirow{2}{*}{\begin{tabular}[c]{@{}l@{}}Language\\ Adaptation\end{tabular}}} & El$\leftrightarrow$En & 1M & \multirow{2}{*}{997} & \multirow{2}{*}{1012} \\
\multicolumn{1}{l|}{} & Sk$\leftrightarrow$En & 1M &  &  \\ \hline
\end{tabular}
\caption{The statics of our datasets. The number in Valid/Test columns denotes the amount of sentence pairs in each domain or translation direction.}
\label{tab:data}
\end{table}

\subsection{Data Preparation}
\textbf{Multilingual Translation Task}  
We test the performance of the model on the original translation task before and after continual learning, to verify whether the methods can preserve the previously learned knowledge.
To achieve this, we use the FLORES-101 test sets~\cite{abs-2106-03193}, which are extracted from English Wikipedia and cover a variety of different topics and domains. 
We test the translation performance of other 49 languages to and from English.
For our method, we also use the FLORES-101 validation sets to compute the empirical Fisher information matrix and search the parameter update regions.

\textbf{Domain Adaptation Task} For the domain adaptation task, we use the data set proposed by~\citet{KoehnK17} to simulate a diverse multi-domain setting. 
The data set includes parallel text in German and English, both of which have already been supported by the mBART50-nn model.
The text is mainly from five diverse domains: IT, Law, Medical, Subtitles, and Koran, which are available in OPUS~\cite{AulamoT19}. 
We use the new split released by~\citet{AharoniG20}, and perform German to English translation (De$\rightarrow$En) for this task.
It should be noted that the De$\rightarrow$En translation task has already been supported by the mBART50-nn model.

\textbf{Language Adaptation Task} For the language adaptation task, we adapt the model to support the Greek$\leftrightarrow$English (El$\leftrightarrow$En) and Slovak$\leftrightarrow$English (Sk$\leftrightarrow$En) translation directions. 
Greek is from an unseen language family and uses an unseen alphabet, which is very different from all the languages supported by mBART50-nn. 
In contrast, Slovak is from the same language family as Czech, which is already supported by the mBART50-nn model, so it is more familiar to the model. 
Therefore, we use them as the new languages because this can simulate most cases where we need language adaptation. 
We use the training data from OPUS-100~\cite{ZhangWTS20} to train the model, and the validation/test sets from FLORES-101 to choose the checkpoint and test the performance. 

Following~\citet{abs-2008-00401}, we use the sentencepiece~\cite{KudoR18} model, which was trained using monolingual data for 100 languages from XLMR, to process all the above data.

\begin{table*}[ht]
\centering
\resizebox{2.\columnwidth}!{
\begin{tabular}{l|ccc|cccccc|c}
\hline
\multirow{2}{*}{Systems} & \multicolumn{3}{c|}{Multilingual Translation} & \multicolumn{6}{c|}{Domain Adaptation} & \multirow{2}{*}{Avg} \\ \cline{2-10}
 & xx$\rightarrow$En & En$\rightarrow$xx & Avg1 & IT & Law & Medical & Subtitles & Koran & Avg2 &  \\ \hline \hline
Scratch & / & / & / & 39.87 & 53.96 & 53.88 & 27.71 & 18.80 & 38.84 & / \\
mBART50-nn & 25.83 & 21.48 & 23.66 & 35.65 & 41.81 & 37.21 & 27.14 & 16.41 & 31.64 & 27.65 \\
Fine-Tuning & 19.27 & 1.64 & 10.46 & 45.99 & 57.02 & 54.71 & 31.98 & 21.58 & 42.26 & 26.36 \\
Mixed-FT & 19.33 & 2.05 & 10.69 & 45.8 & 56.93 & 53.69 & 31.5 & 21.59 & 41.90 & 26.30 \\
\hdashline
KD & 23.85 & 19.55 & 21.7 & 39.71 & 49.13 & 46.64 & 30.58 & 20.2 & 37.25 & 29.48 \\
L2-Reg & 24.03 & 19.85 & 21.94 & 41.03 & 50.88 & 49.19 & 30.1 & 20.5 & 38.34 & 30.14 \\
EWC & 24.19 & 20.29 & 22.24 & 41.02 & 50.25 & 49.2 & 30.59 & 19.68 & 38.15 & 30.19 \\
FT-xattn & 23.35 & 18.44 & 20.90 & 41.44 & 51.43 & 50.03 & 30.64 & 19.95 & 38.70 & 29.80 \\
LFR-CM & \textbf{25.09} & \textbf{20.36} & \textbf{22.73} & 41.73 & 51.24 & 50.28 & 30.96 & 20.98 & 39.04 & 30.89 \\
LFR-OM & 24.78 & 19.48 & 22.13 & \textbf{43.18} & \textbf{52.72} & \textbf{51.44} & \textbf{31.33} & \textbf{21.51} & \textbf{40.04} & \textbf{31.09} \\ \hline
\end{tabular}
}
\caption{The overall BLEU scores of the domain adaptation task. "xx" denotes the languages already supported by mBART50-nn. "Avg1" and "Avg2" denote the average BLEU scores on the multilingual translation task and domain adaptation task, respectively. "Avg" is computed by (Avg1+Avg2)/2 to indicate the balance between the previous and new tasks.  The highest scores among all the continual learning methods are marked in bold.}
\label{tab:res-domain}
\end{table*}

\begin{table*}[ht]
\centering
\resizebox{2\columnwidth}!{
\begin{tabular}{l|ccc|ccccc|c}
\hline
\multirow{2}{*}{Systems} & \multicolumn{3}{c|}{Multilingual Translation} & \multicolumn{5}{c|}{Language Adaptation} & \multirow{2}{*}{Avg} \\ \cline{2-9}
 & xx$\rightarrow$En & En$\rightarrow$xx & Avg1 & El$\rightarrow$En & En$\rightarrow$El & Sk$\rightarrow$En & En$\rightarrow$Sk & Avg2 &  \\ \hline
 \hline
Scratch & / & / & / & 24.93 & 25.39 & 28.17 & 26.59 & 26.27 & / \\
mBART50-nn+LSE & 25.83 & 21.48 & 23.66 & 26.57 & 16.06 & 35.82 & 28.6 & 26.76 & 25.21 \\
Fine-Tuning & 18.37 & 1.15 & 9.76 & 30.59 & 26.67 & 34.96 & 34.06 & 31.57 & 20.67 \\
Mixed-FT & 19.55 & 3.61 & 11.58 & 30.33 & 25.98 & 33.1 & 33.89 & 30.83 & 21.20 \\
\hdashline
L2-Reg & 25.87 & 18.34 & 22.11 & 26.67 & 18.67 & 35.41 & 30.62 & 27.84 & 24.97 \\
EWC & 25.99 & 18.2 & 22.10 & 26.88 & 18.5 & 35.31 & 30.41 & 27.78 & 24.94 \\
FT-xattn & 23.15 & 16.82 & 19.99 & 27.27 & 18.69 & 35.84 & 31.07 & 28.22 & 24.10 \\
LFR-CM & 26.42 & 18.06 & 22.24 & \textbf{28.55} & \textbf{20.81} & \textbf{36.1} & \textbf{31.41} & \textbf{29.22} & \textbf{25.73} \\
LFR-OM & \textbf{26.66} & \textbf{18.68} & \textbf{22.67} & 28.05 & 19.76 & 36.05 & 30.76 & 28.66 & 25.67 \\ \hline
\end{tabular}
}
\caption{The overall BLEU scores of the language adaptation task.}
\label{tab:my-lang}
\end{table*}

\subsection{Systems}
We use the open-source toolkit called {\em Fairseq-py}~\cite{OttEBFGNGA19} as our Transformer system. The following systems can be divided into two categories. 
The first category only focuses on either the previous task or the new task.

\noindent \textbullet \ \textbf{Scratch}: This system is trained from scratch only with the training data from the new translation task.  

\noindent \textbullet \ \textbf{mBART50-nn}~\cite{abs-2008-00401}: The large scale pretrained NMT model. All the following systems are implemented based on this model. 

\noindent \textbullet \ \textbf{mBART50-nn + Language-Specific Embedding (LSE)}~\cite{Berard21}: We insert a new language-specific embedding layer for the new languages and fine-tune these parameters with the new training data for 20k steps. The original parameters are kept fixed during training. 
For the language adaptation task, we use this system as the baseline and other methods are implemented based on this system.

\noindent \textbullet \ \textbf{Fine-tuning}~\cite{luong2015stanford}: This system is trained based on the pretrained model only with the new training data.

\noindent \textbullet \ \textbf{Mixed-FT}: We mix the small-scaled validation sets related to the previous task, i.e., the FLORES-101 validation sets of the languages supported by mBART50-nn, with the new training data to train the model jointly.  
We use the temperature-based sampling function to oversample the validation datasets and the temperature is set as $20$~\cite{abs-1907-05019}.

The second category contains several continual learning methods, which aim to get a good balance between previous and new tasks.

\noindent \textbullet \ \textbf{Knowledge Distillation (KD)}~\cite{dakwale2017fine}: Besides minimizing the training loss of the new task, this method also minimizes the cross-entropy between the output distribution of the mBART50-nn model and the network. The final objective is:
\begin{equation}
    \mathcal{L}_{\mathrm{KD}}(\theta) = \mathcal{L}_{CE}(\theta) + \alpha KL(p_{\theta_0}||p_{\theta}),
\end{equation}
where $\alpha$ is the hyper-parameter which controls the contribution of the two parts. 

\noindent \textbullet \ \textbf{L2-Reg}~\cite{BaroneHGS17}: This method adds a L2-norm regularizations on the parameters:
\begin{equation}
    \mathcal{L}_{\mathrm{L2}}(\theta) = \mathcal{L}_{CE}(\theta) + \frac{\alpha}{M} \sum_{i=0}^{M} (\theta_i - \theta_{0,i})^2,
\end{equation}
where $i$ denotes the $i$-th parameter.

\noindent \textbullet \ \textbf{EWC}~\cite{ThompsonGKDK19}: This method models the importance of the parameters with Fisher information matrix and puts more constrains on the important parameters to let them stay close to the original values. The training objective is:
\begin{equation}
    \mathcal{L}_{\mathrm{EWC}}(\theta) = \mathcal{L}_{CE}(\theta) + \frac{\alpha}{M} \sum_{i=0}^{M} F_i (\theta_i - \theta_{0,i})^2,
\end{equation}
where $F_i$ denotes the modeled importance for the $i$-th parameter.

\noindent \textbullet \ \textbf{FT-xattn}~\cite{Gheini0M21}: This method only fine-tunes the cross-attention parameters, which can also mitigate the forgetting problem. 

\noindent \textbullet \ \textbf{LFR-CM}: This system is implemented as the curvature-based method. We set $\rho\%$ as $75\%$ and $\lambda$ as $0.1$ in the main experiments. We put more results about the hyperparameters in the next section and Appendix. 

\noindent \textbullet \ \textbf{LFR-OM}: This system is implemented as the output-based method. We set the hyper-parameter as $1$. Other training details are listed below.

\textbf{Training Details} We set dropout as $0.3$ and attention-dropout as $0.1$. We employ the Adam optimizer with $\beta_1 = 0.9$ and $\beta_2 = 0.98$. 
We use the inverse square root learning scheduler and set the $warmup\_steps=4000$. 
For the curvature-based method (LFR-CM), we set $\rho\%$ as $75\%$ and $\lambda$ as $0.2$.
For the output-based method (LFR-OM), we set $lr=2e-4$ for the domain adaptation task, $lr=4e-5$ for the language adaptation, and train the model $5$k steps to search the parameter update region. During continual learning, we set $lr=5e-4$ and train the model $30$k steps for the domain adaptation task; we set $lr=5e-5$ and train the model $50$k steps for the language adaptation task. We fix all the norm layers of the model in both of the two tasks. In the language adaptation task, we also fix the original and new embedding layers, which we find can also help alleviate the forgetting problem. All the systems are trained on 8 A100 GPUs with the update frequency 2. The max token is $1024$ for each GPU. Besides, we use beam search with the size of 4 and length penalty as 1 during decoding.

\begin{figure}[t]
    \centering
    \subfigure[Domain Adaptation Task]{
        \includegraphics[width=0.9\columnwidth]{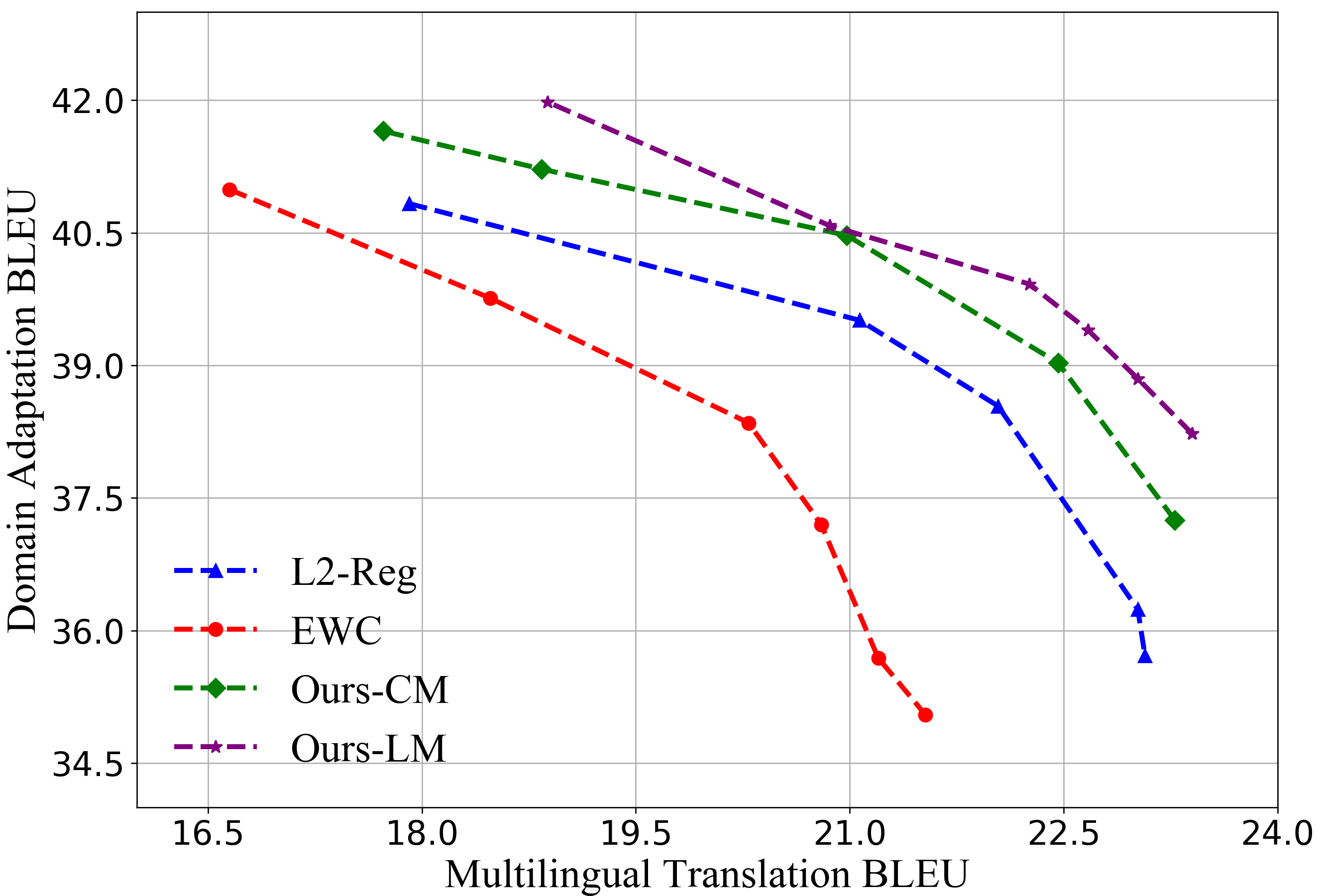}
    }
    \subfigure[Language Adaptation Task]{
        \includegraphics[width=0.9\columnwidth]{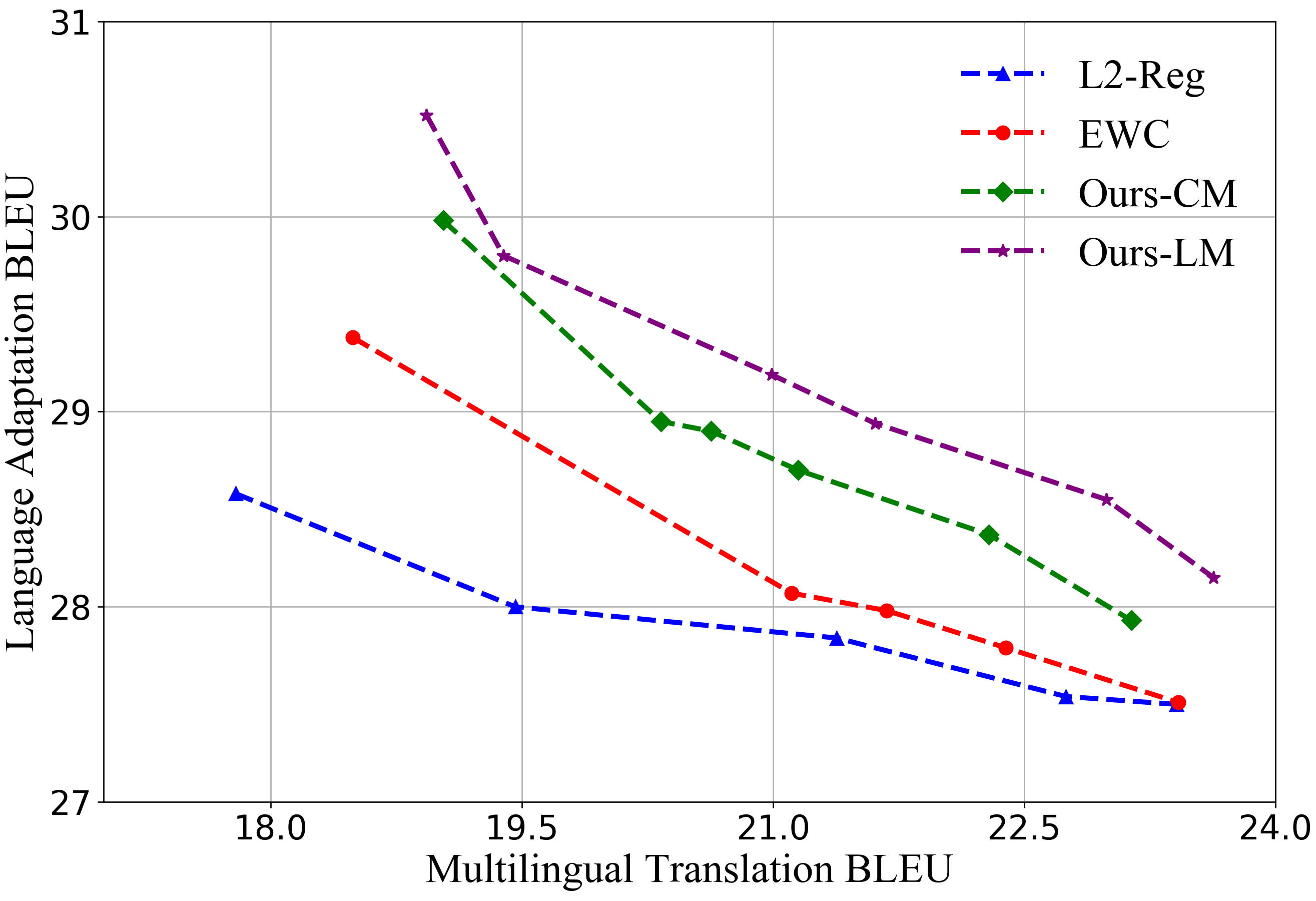}
    }
    \caption{The performance trade-off with different hyper-parameters. The x-axis and y-axis denote the BLEU of the previous and new tasks, respectively. The closer the point is to the upper right corner, the better the performance.}
    \label{fig:hyper}
\end{figure}

\subsection{Main Results}
The final translation is evaluated using the $4$-gram case-sensitive BLEU~\cite{PapineniRWZ02} with the {\em SacreBLEU} tool~\cite{post-2018-call}. Following~\citet{abs-2106-03193}, we report the SentencePiece BLEU which uses a SentencePiece tokenizer (SPM) with 256K tokens and then BLEU score is computed on the sentence-piece tokenized text~\footnote{BLEU+case.mixed+numrefs.1+smooth.exp\\+tok.spm+version.1.5.1}.

The main results of the domain adaptation task are in Table~\ref{tab:res-domain}. 
Compared with the Scratch system, Fine-Tuning can greatly improve the performance of domain adaptation tasks, but it also suffers from catastrophic forgetting on the multilingual translation task. 
Besides, we observe that the forgetting problem in En$\rightarrow$xx directions is more severer. 
After analyzing the model output, we find that the model cannot output other languages except for English for all previous translation directions.
This may be because the target language of the domain adaptation task is only English. 
Among all the continual learning methods, our method can get the best overall performance. 
The main results of the language adaptation task are in Table~\ref{tab:my-lang}. 
By adding and updating the new language-specific embeddings (mBART50-nn+LSE), we can achieve good results except for the En$\rightarrow$El directions.
Greek is quite different from the previous languages, so it is more difficult for the model to learn it as the target language only by the existing knowledge.
Just like the domain adaptation task, our method outperforms other continual learning methods.

\section{Analysis}

\subsection{Effects of Different Hyper-parameters}
For the regularization-based methods, the hyper-parameter $\alpha$ controls the performance trade-off between the previous and new tasks.
The larger the hyper-parameters $\alpha$ is, the less decline of the BLEU on the original task will be, and the less improvement of the new task performance will be. 
As for our method, 
the proportion of model parameters to be pruned has a similar effect.  
erasing more neurons will bring better results in the new task, but will also lead to more degradation in the original task. 
To better show the full performance trade-off, we conduct experiments with different hyper-parameters. We compare our method with the L2-Reg and EWC systems.
For the L2-Reg and EWC method, we vary $\alpha$ from $0.001$ to $1$. 
For the curvature-based method, we fix $\rho\%$ as $75\%$ and vary $\lambda$ from $0.1$ to $1$. For the learning-based method, we vary the learning rate from $1e-5$ to $1e-4$, because we find that adjusting the learning rate is more effective than changing the hyper-parameter. The detailed settings of the hyper-parameters are put in the Appendix. 
The results are shown in Figure~\ref{fig:hyper}.
It shows that our method outperforms L2-Reg and EWC at all the operating points significantly. Besides, it also shows that the output-based method is better than the curvature-based method.


\begin{table}[]
\centering
\resizebox{\columnwidth}!{
\begin{tabular}{l|ccc}
\hline
System & \begin{tabular}[c]{@{}c@{}}\{Zh,Fr,De\}\\ $\leftrightarrow$EN Avg.\end{tabular} & \begin{tabular}[c]{@{}c@{}}\{El, Sk\}\\ $\leftrightarrow$En Avg.\end{tabular} & \begin{tabular}[c]{@{}c@{}}Zero\\ Avg.\end{tabular} \\ \hline \hline
mBart50+LSE & 35.24 & 26.76 & 5.80 \\
Scratch & 26.68 & 21.82 & 6.81 \\
Fine-tuning & 32.30 & 28.51 & 5.07 \\
\hdashline
L2-Reg & 30.09 & 22.53 & 2.40 \\
EWC & 26.32 & 21.02 & 3.06 \\
LFR-CM & 34.53 & \textbf{29.83} & \textbf{8.55} \\
LFR-OM & \textbf{34.88} & 28.94 & 8.12 \\ \hline
\end{tabular}
}
\caption{BLEU of the mixed-training experiments. "Zero Avg." denotes the average BLEU of \{Zh, Fr, De\}$\leftrightarrow$\{El, Sk\}.}
\label{tab:mix}
\end{table}

\subsection{Mixed-Training with Previous Data}
In this work, we try to preserve all the previously learned knowledge and consider the situation that performing continual learning without access to any previous training data.
But in practice, it is more likely that we just want to preserve some specific knowledge, e.g., for some specific languages, and we have some training data from or related to the previous training task.
To prove the effectiveness of our method under this scenario, we conduct further experiments on the language adaptation task. 
We choose Chinese (Zh$\leftrightarrow$En), French (Fr$\leftrightarrow$En), and German (De$\leftrightarrow$En) from the languages supported by mBART50-nn as our target languages, on which we want to retain the translation performance. 
Then we collect the corresponding training data from the OPUS-100 dataset, which is much smaller in quantity than the previous training data of mBART50-nn.
Last we continually train the model with the mixed data.
We test the BLEU scores on the supervised directions and the zero-shot directions between the previous languages (Zh, Fr, De) and the new languages (El, Sk).
The results are put in the Table~\ref{tab:mix}, and the detailed hyper-parameter settings are put in the Appendix. 
We find that the regularization-based method fails to deal with this scenario, and they are even worse than the vanilla Fine-tuning method, which indicates that the soft constraints they put on the parameters are harmful to the model when some previous data is available. 
Compared with the fine-tuning method, our method can further reduce the forgetting problem with the previous training data and achieve better overall performance.

\begin{table}[t]
\centering
\resizebox{\columnwidth}!{
\begin{tabular}{l|cccc}
\hline
System & xx$\rightarrow$En & En$\rightarrow$xx & El$\leftrightarrow$En & Sk$\leftrightarrow$En \\ \hline \hline
Fine-tuning & 23.91 & 1.99 & 14.43 & 34.5 \\
\hdashline
L2-Reg & 25.95 & 20.53 & 21.32 & 32.58 \\
EWC & 26.02 & 20.46 & 21.21 & 32.04 \\
LFR-CM & 26.64 & 20.83 & 21.85 & 32.84 \\
LFR-OM & \textbf{26.69} & \textbf{21.39} & \textbf{22.03} & \textbf{33.03} \\ \hline
\end{tabular}
}
\caption{BLEU of the sequential language adaptation experiments.}
\label{tab:sla}
\end{table}

\subsection{Sequential Language Adaptation}
In this experiment, we perform the language adaptation task in the sequential training scenario. 
We first train the model with the EL$\leftrightarrow$En data and then with the Sk$\leftrightarrow$En data. 
For the L2-Reg and EWC methods, we use the model after training with the EL$\leftrightarrow$En data as the pretrained model for the Sk$\leftrightarrow$En task to compute the regularization loss.
We recompute the Fisher information matrix for the EWC and LFR-CM method after the El$\leftrightarrow$En training stage.
Besides, we also recompute the update regions after the training of El$\leftrightarrow$En for the Sk$\leftrightarrow$En task.
The detailed hyper-parameter settings are put in the Appendix.
We report the final results in Table~\ref{tab:sla}. The results show that our method still outperforms other methods under this scenario. 

\section{Related Work}
Recent work on continual learning of NMT can be divided into three categories: data memory based method, task-specific module based method, and regularization based method.

\textbf{Data Memory Based Method} The data memory based methods need to retain part or all of the training data of the previous task. \citet{chu2017empirical} fine-tune the pretrained model with the mix of the previous and new data. \citet{BapnaF19} propose an n-gram level retrieval approach to find the useful context to help translation. \citet{XuCS20} propose to use the similar translations from translation memory to boost the performance. 
\citet{LiuWF21} utilize a bilingual dictionary to generate mixed-language sentences. 
Compared with these methods, our method doesn't need the previous data and thus is more flexible in practice.


\textbf{Task-specific Module Based Method} The task-specific module based methods need to assign the model parameters to different tasks. \citet{BapnaF19} injects domain-specific adapter modules into each layer of the pretrained model, then they fix the model and only train the new adapters. Based on this, here are also some work~\cite{ZengSWLXYZ18,ZengLSGLYL19,GuFL19,CaoWCW21} that adds different kinds of task-specific modules to the model. Besides, some work tries to divide the model into different parts based on the parameter importance on each task~\cite{GuFX21,liang2020finding}. Although they don't increase the model, these methods actually divide the model into different task-specific parts, which makes the model need to know which task the input comes from. Compared with these methods, our method doesn't introduce model separation explicitly. 

\textbf{Regularization Based Method} The regularization based methods mitigate forgetting by introducing an additional penalty term in the learning objective. \citet{KhayrallahTDK18} and \citet{ThompsonGKDK19} add regularization terms to let the model parameters stay close to their original values. \citet{dakwale2017fine} minimize the cross-entropy between the output distribution of the pretrained model and the fine-tuned model. Different from the above work, our method constrains the model parameters within low forgetting risk regions to avoid the forgetting problem.


Besides, our work is also related to unstructured model pruning~\cite{see-etal-2016-compression,ZhuG18,FrankleC19}, because we both aim to search and operate on the unimportant parameters. The difference is that our method is to find an update region around the initial parameters, while model pruning should directly remove some parameters, that is, set them to zero.

\section{Conclusion}
In this work, we propose a continual learning method for NMT by constraining model parameters within low forgetting risk regions. We propose two methods to find such regions, the first is based on the curvature of the loss function and the second is based on the model output of the previous task. Then we can continually train the model parameters within this region. 
The experimental results on the domain adaptation and language adaptation tasks prove that our method can achieve significant improvements over several strong baselines. 

\section*{Limitations}
Because our method does not introduce additional parameters or use the previous data, the overall performance of our method is weaker than the data memory based methods and task-specific module based methods. 
It is difficult for our method to achieve the same performance as the Fine-tuning method in the new task without causing catastrophic forgetting.

\section*{Acknowledgements}
We thank all the anonymous reviewers for their insightful and valuable comments.


\bibliography{anthology,custom}

\begin{thebibliography}{41}
\expandafter\ifx\csname natexlab\endcsname\relax\def\natexlab#1{#1}\fi

\bibitem[{Aharoni and Goldberg(2020)}]{AharoniG20}
Roee Aharoni and Yoav Goldberg. 2020.
\newblock \href {https://doi.org/10.18653/v1/2020.acl-main.692} {Unsupervised
  domain clusters in pretrained language models}.
\newblock In \emph{Proceedings of the 58th Annual Meeting of the Association
  for Computational Linguistics, {ACL} 2020, Online, July 5-10, 2020}, pages
  7747--7763.

\bibitem[{Arivazhagan et~al.(2019)Arivazhagan, Bapna, Firat, Lepikhin, Johnson,
  Krikun, Chen, Cao, Foster, Cherry, Macherey, Chen, and Wu}]{abs-1907-05019}
Naveen Arivazhagan, Ankur Bapna, Orhan Firat, Dmitry Lepikhin, Melvin Johnson,
  Maxim Krikun, Mia~Xu Chen, Yuan Cao, George~F. Foster, Colin Cherry, Wolfgang
  Macherey, Zhifeng Chen, and Yonghui Wu. 2019.
\newblock \href {http://arxiv.org/abs/1907.05019} {Massively multilingual
  neural machine translation in the wild: Findings and challenges}.
\newblock \emph{CoRR}, abs/1907.05019.

\bibitem[{Aulamo and Tiedemann(2019)}]{AulamoT19}
Mikko Aulamo and J{\"{o}}rg Tiedemann. 2019.
\newblock \href {https://aclanthology.org/W19-6146/} {The {OPUS} resource
  repository: An open package for creating parallel corpora and machine
  translation services}.
\newblock In \emph{Proceedings of the 22nd Nordic Conference on Computational
  Linguistics, NoDaLiDa 2019, Turku, Finland, September 30 - October 2, 2019},
  pages 389--394.

\bibitem[{Bapna and Firat(2019)}]{BapnaF19}
Ankur Bapna and Orhan Firat. 2019.
\newblock \href {https://doi.org/10.18653/v1/n19-1191} {Non-parametric
  adaptation for neural machine translation}.
\newblock In \emph{Proceedings of the 2019 Conference of the North American
  Chapter of the Association for Computational Linguistics: Human Language
  Technologies, {NAACL-HLT} 2019, Minneapolis, MN, USA, June 2-7, 2019, Volume
  1 (Long and Short Papers)}, pages 1921--1931.

\bibitem[{Barone et~al.(2017)Barone, Haddow, Germann, and
  Sennrich}]{BaroneHGS17}
Antonio Valerio~Miceli Barone, Barry Haddow, Ulrich Germann, and Rico Sennrich.
  2017.
\newblock \href {https://doi.org/10.18653/v1/d17-1156} {Regularization
  techniques for fine-tuning in neural machine translation}.
\newblock In \emph{Proceedings of the 2017 Conference on Empirical Methods in
  Natural Language Processing, {EMNLP} 2017, Copenhagen, Denmark, September
  9-11, 2017}, pages 1489--1494.

\bibitem[{Berard(2021)}]{Berard21}
Alexandre Berard. 2021.
\newblock \href {https://aclanthology.org/2021.wmt-1.62} {Continual learning in
  multilingual {NMT} via language-specific embeddings}.
\newblock In \emph{Proceedings of the Sixth Conference on Machine Translation,
  WMT@EMNLP 2021, Online Event, November 10-11, 2021}, pages 542--565.

\bibitem[{Cao et~al.(2021)Cao, Wei, Chen, and Wan}]{CaoWCW21}
Yue Cao, Hao{-}Ran Wei, Boxing Chen, and Xiaojun Wan. 2021.
\newblock \href {https://doi.org/10.18653/v1/2021.naacl-main.310} {Continual
  learning for neural machine translation}.
\newblock In \emph{Proceedings of the 2021 Conference of the North American
  Chapter of the Association for Computational Linguistics: Human Language
  Technologies, {NAACL-HLT} 2021, Online, June 6-11, 2021}, pages 3964--3974.

\bibitem[{Chu et~al.(2017)Chu, Dabre, and Kurohashi}]{chu2017empirical}
Chenhui Chu, Raj Dabre, and Sadao Kurohashi. 2017.
\newblock An empirical comparison of simple domain adaptation methods for
  neural machine translation.
\newblock \emph{arXiv preprint arXiv:1701.03214}.

\bibitem[{Dakwale and Monz(2017)}]{dakwale2017fine}
Praveen Dakwale and Christof Monz. 2017.
\newblock Fine-tuning for neural machine translation with limited degradation
  across in-and out-of-domain data.
\newblock \emph{Proceedings of the XVI Machine Translation Summit}, page 117.

\bibitem[{Escolano et~al.(2021)Escolano, Costa{-}juss{\`{a}}, and
  Fonollosa}]{EscolanoCF21}
Carlos Escolano, Marta~R. Costa{-}juss{\`{a}}, and Jos{\'{e}} A.~R. Fonollosa.
  2021.
\newblock \href {https://doi.org/10.1002/asi.24395} {From bilingual to
  multilingual neural-based machine translation by incremental training}.
\newblock \emph{J. Assoc. Inf. Sci. Technol.}, 72(2):190--203.

\bibitem[{Fan et~al.(2021)Fan, Bhosale, Schwenk, Ma, El{-}Kishky, Goyal,
  Baines, Celebi, Wenzek, Chaudhary, Goyal, Birch, Liptchinsky, Edunov, Auli,
  and Joulin}]{FanBSMEGBCWCGBL21}
Angela Fan, Shruti Bhosale, Holger Schwenk, Zhiyi Ma, Ahmed El{-}Kishky,
  Siddharth Goyal, Mandeep Baines, Onur Celebi, Guillaume Wenzek, Vishrav
  Chaudhary, Naman Goyal, Tom Birch, Vitaliy Liptchinsky, Sergey Edunov,
  Michael Auli, and Armand Joulin. 2021.
\newblock \href {http://jmlr.org/papers/v22/20-1307.html} {Beyond
  english-centric multilingual machine translation}.
\newblock \emph{J. Mach. Learn. Res.}, 22:107:1--107:48.

\bibitem[{Frankle and Carbin(2019)}]{FrankleC19}
Jonathan Frankle and Michael Carbin. 2019.
\newblock \href {https://openreview.net/forum?id=rJl-b3RcF7} {The lottery
  ticket hypothesis: Finding sparse, trainable neural networks}.
\newblock In \emph{7th International Conference on Learning Representations,
  {ICLR} 2019, New Orleans, LA, USA, May 6-9, 2019}. OpenReview.net.

\bibitem[{Gheini et~al.(2021)Gheini, Ren, and May}]{Gheini0M21}
Mozhdeh Gheini, Xiang Ren, and Jonathan May. 2021.
\newblock \href {https://doi.org/10.18653/v1/2021.emnlp-main.132}
  {Cross-attention is all you need: Adapting pretrained transformers for
  machine translation}.
\newblock In \emph{Proceedings of the 2021 Conference on Empirical Methods in
  Natural Language Processing, {EMNLP} 2021, Virtual Event / Punta Cana,
  Dominican Republic, 7-11 November, 2021}, pages 1754--1765.

\bibitem[{Goyal et~al.(2021)Goyal, Gao, Chaudhary, Chen, Wenzek, Ju, Krishnan,
  Ranzato, Guzm{\'{a}}n, and Fan}]{abs-2106-03193}
Naman Goyal, Cynthia Gao, Vishrav Chaudhary, Peng{-}Jen Chen, Guillaume Wenzek,
  Da~Ju, Sanjana Krishnan, Marc'Aurelio Ranzato, Francisco Guzm{\'{a}}n, and
  Angela Fan. 2021.
\newblock \href {http://arxiv.org/abs/2106.03193} {The {FLORES-101} evaluation
  benchmark for low-resource and multilingual machine translation}.
\newblock \emph{CoRR}, abs/2106.03193.

\bibitem[{Gu and Feng(2020)}]{gu-feng-2020-investigating}
Shuhao Gu and Yang Feng. 2020.
\newblock \href {https://doi.org/10.18653/v1/2020.coling-main.381}
  {Investigating catastrophic forgetting during continual training for neural
  machine translation}.
\newblock In \emph{Proceedings of the 28th International Conference on
  Computational Linguistics}, pages 4315--4326, Barcelona, Spain (Online).
  International Committee on Computational Linguistics.

\bibitem[{Gu et~al.(2019)Gu, Feng, and Liu}]{GuFL19}
Shuhao Gu, Yang Feng, and Qun Liu. 2019.
\newblock Improving domain adaptation translation with domain invariant and
  specific information.
\newblock In \emph{Proceedings of the 2019 Conference of the North American
  Chapter of the Association for Computational Linguistics: Human Language
  Technologies, {NAACL-HLT}}, pages 3081--3091.

\bibitem[{Gu et~al.(2021)Gu, Feng, and Xie}]{GuFX21}
Shuhao Gu, Yang Feng, and Wanying Xie. 2021.
\newblock \href {https://doi.org/10.18653/v1/2021.naacl-main.308}
  {Pruning-then-expanding model for domain adaptation of neural machine
  translation}.
\newblock In \emph{Proceedings of the 2021 Conference of the North American
  Chapter of the Association for Computational Linguistics: Human Language
  Technologies, {NAACL-HLT} 2021, Online, June 6-11, 2021}, pages 3942--3952.

\bibitem[{Hoefler et~al.(2021)Hoefler, Alistarh, Ben{-}Nun, Dryden, and
  Peste}]{HoeflerABDP21}
Torsten Hoefler, Dan Alistarh, Tal Ben{-}Nun, Nikoli Dryden, and Alexandra
  Peste. 2021.
\newblock \href {http://jmlr.org/papers/v22/21-0366.html} {Sparsity in deep
  learning: Pruning and growth for efficient inference and training in neural
  networks}.
\newblock \emph{J. Mach. Learn. Res.}, 22:241:1--241:124.

\bibitem[{Johnson et~al.(2017)Johnson, Schuster, Le, Krikun, Wu, Chen, Thorat,
  Vi{\'{e}}gas, Wattenberg, Corrado, Hughes, and Dean}]{JohnsonSLKWCTVW17}
Melvin Johnson, Mike Schuster, Quoc~V. Le, Maxim Krikun, Yonghui Wu, Zhifeng
  Chen, Nikhil Thorat, Fernanda~B. Vi{\'{e}}gas, Martin Wattenberg, Greg
  Corrado, Macduff Hughes, and Jeffrey Dean. 2017.
\newblock \href {https://doi.org/10.1162/tacl\_a\_00065} {Google's multilingual
  neural machine translation system: Enabling zero-shot translation}.
\newblock \emph{Trans. Assoc. Comput. Linguistics}, 5:339--351.

\bibitem[{Khayrallah et~al.(2018)Khayrallah, Thompson, Duh, and
  Koehn}]{KhayrallahTDK18}
Huda Khayrallah, Brian Thompson, Kevin Duh, and Philipp Koehn. 2018.
\newblock \href {https://doi.org/10.18653/v1/w18-2705} {Regularized training
  objective for continued training for domain adaptation in neural machine
  translation}.
\newblock In \emph{Proceedings of the 2nd Workshop on Neural Machine
  Translation and Generation, NMT@ACL 2018, Melbourne, Australia, July 20,
  2018}, pages 36--44.

\bibitem[{Kim and Rush(2016)}]{KimR16}
Yoon Kim and Alexander~M. Rush. 2016.
\newblock \href {https://doi.org/10.18653/v1/d16-1139} {Sequence-level
  knowledge distillation}.
\newblock In \emph{Proceedings of the 2016 Conference on Empirical Methods in
  Natural Language Processing, {EMNLP} 2016, Austin, Texas, USA, November 1-4,
  2016}, pages 1317--1327.

\bibitem[{Ko et~al.(2021)Ko, El{-}Kishky, Renduchintala, Chaudhary, Goyal,
  Guzm{\'{a}}n, Fung, Koehn, and Diab}]{KoERCGGFKD20}
Wei{-}Jen Ko, Ahmed El{-}Kishky, Adithya Renduchintala, Vishrav Chaudhary,
  Naman Goyal, Francisco Guzm{\'{a}}n, Pascale Fung, Philipp Koehn, and Mona~T.
  Diab. 2021.
\newblock \href {https://doi.org/10.18653/v1/2021.acl-long.66} {Adapting
  high-resource {NMT} models to translate low-resource related languages
  without parallel data}.
\newblock In \emph{Proceedings of the 59th Annual Meeting of the Association
  for Computational Linguistics and the 11th International Joint Conference on
  Natural Language Processing, {ACL/IJCNLP} 2021, (Volume 1: Long Papers),
  Virtual Event, August 1-6, 2021}, pages 802--812.

\bibitem[{Koehn and Knowles(2017)}]{KoehnK17}
Philipp Koehn and Rebecca Knowles. 2017.
\newblock \href {https://doi.org/10.18653/v1/w17-3204} {Six challenges for
  neural machine translation}.
\newblock In \emph{Proceedings of the First Workshop on Neural Machine
  Translation, NMT@ACL 2017, Vancouver, Canada, August 4, 2017}, pages 28--39.

\bibitem[{Kudo and Richardson(2018)}]{KudoR18}
Taku Kudo and John Richardson. 2018.
\newblock \href {https://doi.org/10.18653/v1/d18-2012} {Sentencepiece: {A}
  simple and language independent subword tokenizer and detokenizer for neural
  text processing}.
\newblock In \emph{Proceedings of the 2018 Conference on Empirical Methods in
  Natural Language Processing, {EMNLP} 2018: System Demonstrations, Brussels,
  Belgium, October 31 - November 4, 2018}, pages 66--71.

\bibitem[{Liang et~al.(2020)Liang, Zhao, Wang, Qiu, and Li}]{liang2020finding}
Jianze Liang, Chengqi Zhao, Mingxuan Wang, Xipeng Qiu, and Lei Li. 2020.
\newblock Finding sparse structure for domain specific neural machine
  translation.
\newblock \emph{AAAI 2021}.

\bibitem[{Liu et~al.(2020)Liu, Gu, Goyal, Li, Edunov, Ghazvininejad, Lewis, and
  Zettlemoyer}]{LiuGGLEGLZ20}
Yinhan Liu, Jiatao Gu, Naman Goyal, Xian Li, Sergey Edunov, Marjan
  Ghazvininejad, Mike Lewis, and Luke Zettlemoyer. 2020.
\newblock \href {https://transacl.org/ojs/index.php/tacl/article/view/2107}
  {Multilingual denoising pre-training for neural machine translation}.
\newblock \emph{Trans. Assoc. Comput. Linguistics}, 8:726--742.

\bibitem[{Liu et~al.(2021)Liu, Winata, and Fung}]{LiuWF21}
Zihan Liu, Genta~Indra Winata, and Pascale Fung. 2021.
\newblock \href {https://doi.org/10.18653/v1/2021.findings-acl.239} {Continual
  mixed-language pre-training for extremely low-resource neural machine
  translation}.
\newblock In \emph{Findings of the Association for Computational Linguistics:
  {ACL/IJCNLP} 2021, Online Event, August 1-6, 2021}, pages 2706--2718.

\bibitem[{Luong and Manning(2015)}]{luong2015stanford}
Minh-Thang Luong and Christopher~D Manning. 2015.
\newblock Stanford neural machine translation systems for spoken language
  domains.
\newblock In \emph{Proceedings of the International Workshop on Spoken Language
  Translation}, pages 76--79.

\bibitem[{Ly et~al.(2017)Ly, Marsman, Verhagen, Grasman, and
  Wagenmakers}]{ly2017tutorial}
Alexander Ly, Maarten Marsman, Josine Verhagen, Raoul~PPP Grasman, and Eric-Jan
  Wagenmakers. 2017.
\newblock A tutorial on fisher information.
\newblock \emph{Journal of Mathematical Psychology}, 80:40--55.

\bibitem[{Ott et~al.(2019)Ott, Edunov, Baevski, Fan, Gross, Ng, Grangier, and
  Auli}]{OttEBFGNGA19}
Myle Ott, Sergey Edunov, Alexei Baevski, Angela Fan, Sam Gross, Nathan Ng,
  David Grangier, and Michael Auli. 2019.
\newblock \href {https://doi.org/10.18653/v1/n19-4009} {fairseq: {A} fast,
  extensible toolkit for sequence modeling}.
\newblock In \emph{Proceedings of the 2019 Conference of the North American
  Chapter of the Association for Computational Linguistics: Human Language
  Technologies, {NAACL-HLT} 2019, Minneapolis, MN, USA, June 2-7, 2019,
  Demonstrations}, pages 48--53.

\bibitem[{Papineni et~al.(2002)Papineni, Roukos, Ward, and Zhu}]{PapineniRWZ02}
Kishore Papineni, Salim Roukos, Todd Ward, and Wei{-}Jing Zhu. 2002.
\newblock \href {https://doi.org/10.3115/1073083.1073135} {Bleu: a method for
  automatic evaluation of machine translation}.
\newblock In \emph{Proceedings of the 40th Annual Meeting of the Association
  for Computational Linguistics, July 6-12, 2002, Philadelphia, PA, {USA}},
  pages 311--318.

\bibitem[{Post(2018)}]{post-2018-call}
Matt Post. 2018.
\newblock \href {https://www.aclweb.org/anthology/W18-6319} {A call for clarity
  in reporting {BLEU} scores}.
\newblock In \emph{Proceedings of the Third Conference on Machine Translation:
  Research Papers}, pages 186--191, Belgium, Brussels. Association for
  Computational Linguistics.

\bibitem[{See et~al.(2016)See, Luong, and Manning}]{see-etal-2016-compression}
Abigail See, Minh-Thang Luong, and Christopher~D. Manning. 2016.
\newblock \href {https://doi.org/10.18653/v1/K16-1029} {Compression of neural
  machine translation models via pruning}.
\newblock In \emph{Proceedings of the 20th {SIGNLL} Conference on Computational
  Natural Language Learning}, pages 291--301, Berlin, Germany. Association for
  Computational Linguistics.

\bibitem[{Tang et~al.(2020)Tang, Tran, Li, Chen, Goyal, Chaudhary, Gu, and
  Fan}]{abs-2008-00401}
Yuqing Tang, Chau Tran, Xian Li, Peng{-}Jen Chen, Naman Goyal, Vishrav
  Chaudhary, Jiatao Gu, and Angela Fan. 2020.
\newblock \href {http://arxiv.org/abs/2008.00401} {Multilingual translation
  with extensible multilingual pretraining and finetuning}.
\newblock \emph{CoRR}, abs/2008.00401.

\bibitem[{Thompson et~al.(2019)Thompson, Gwinnup, Khayrallah, Duh, and
  Koehn}]{ThompsonGKDK19}
Brian Thompson, Jeremy Gwinnup, Huda Khayrallah, Kevin Duh, and Philipp Koehn.
  2019.
\newblock Overcoming catastrophic forgetting during domain adaptation of neural
  machine translation.
\newblock In \emph{NAACL-HLT 2019}, pages 2062--2068.

\bibitem[{Vaswani et~al.(2017)Vaswani, Shazeer, Parmar, Uszkoreit, Jones,
  Gomez, Kaiser, and Polosukhin}]{VaswaniSPUJGKP17}
Ashish Vaswani, Noam Shazeer, Niki Parmar, Jakob Uszkoreit, Llion Jones,
  Aidan~N. Gomez, Lukasz Kaiser, and Illia Polosukhin. 2017.
\newblock Attention is all you need.
\newblock In \emph{Advances in Neural Information Processing Systems 30: Annual
  Conference on Neural Information Processing Systems}, pages 5998--6008.

\bibitem[{Xu et~al.(2020)Xu, Crego, and Senellart}]{XuCS20}
Jitao Xu, Josep~Maria Crego, and Jean Senellart. 2020.
\newblock \href {https://doi.org/10.18653/v1/2020.acl-main.144} {Boosting
  neural machine translation with similar translations}.
\newblock In \emph{Proceedings of the 58th Annual Meeting of the Association
  for Computational Linguistics, {ACL} 2020, Online, July 5-10, 2020}, pages
  1580--1590.

\bibitem[{Zeng et~al.(2019)Zeng, Liu, Su, Ge, Lu, Yin, and Luo}]{ZengLSGLYL19}
Jiali Zeng, Yang Liu, Jinsong Su, Yubin Ge, Yaojie Lu, Yongjing Yin, and Jiebo
  Luo. 2019.
\newblock \href {https://doi.org/10.18653/v1/D19-1078} {Iterative dual domain
  adaptation for neural machine translation}.
\newblock In \emph{Proceedings of the 2019 Conference on Empirical Methods in
  Natural Language Processing and the 9th International Joint Conference on
  Natural Language Processing, {EMNLP-IJCNLP} 2019, Hong Kong, China, November
  3-7, 2019}, pages 845--855.

\bibitem[{Zeng et~al.(2018)Zeng, Su, Wen, Liu, Xie, Yin, and
  Zhao}]{ZengSWLXYZ18}
Jiali Zeng, Jinsong Su, Huating Wen, Yang Liu, Jun Xie, Yongjing Yin, and
  Jianqiang Zhao. 2018.
\newblock \href {https://doi.org/10.18653/v1/d18-1041} {Multi-domain neural
  machine translation with word-level domain context discrimination}.
\newblock In \emph{Proceedings of the 2018 Conference on Empirical Methods in
  Natural Language Processing, Brussels, Belgium, October 31 - November 4,
  2018}, pages 447--457.

\bibitem[{Zhang et~al.(2020)Zhang, Williams, Titov, and Sennrich}]{ZhangWTS20}
Biao Zhang, Philip Williams, Ivan Titov, and Rico Sennrich. 2020.
\newblock \href {https://doi.org/10.18653/v1/2020.acl-main.148} {Improving
  massively multilingual neural machine translation and zero-shot translation}.
\newblock In \emph{Proceedings of the 58th Annual Meeting of the Association
  for Computational Linguistics, {ACL} 2020, Online, July 5-10, 2020}, pages
  1628--1639.

\bibitem[{Zhu and Gupta(2018)}]{ZhuG18}
Michael Zhu and Suyog Gupta. 2018.
\newblock \href {https://openreview.net/forum?id=Sy1iIDkPM} {To prune, or not
  to prune: Exploring the efficacy of pruning for model compression}.
\newblock In \emph{6th International Conference on Learning Representations,
  {ICLR} 2018, Vancouver, BC, Canada, April 30 - May 3, 2018, Workshop Track
  Proceedings}. OpenReview.net.

\end{thebibliography}
\bibliographystyle{acl_natbib}

\appendix

\section{Appendix}
\label{sec:appendix}
\subsection{The mBART50-nn Model}
The mBart50-nn model is a many-to-many multilingual NMT model which can support the translation between 50 different languages. 
The encoder and decoder of mBART50-nn have 12 layers. The dimensionality of the model is set as 1024, while the inner-layer of the feed-forward network has dimensionality as 4096. The attention module has 16 attention heads both in the encoder and decoder. Besides, the model has a shared source-target vocabulary of about 250k tokens, and the model uses learned positional embeddings with the max token length set as 1024. 

\subsection{Hyper-parameter Settings for the Analysis}
In this section, we report the detailed hyper-parameter settings in our experiments. 

For the main experiments, we set $\alpha$ as $1$ for the KD method, $0.01$ for the L2-Reg method, and $0.05$ for the EWC method. We set $\rho\%$ as $75\%$ and $\lambda$ as $0.1$ for the LFR-CM method, and $\alpha$ as $1$ for the LFR-OM method. 

For the experiments studying the effects of different hyper-parameters (Section 5.1), we tried the following hyper-parameters:
\begin{itemize}
    \item L2-Reg ($\alpha$): 0.001, 0.005, 0.01, 0.05, 0.1, 0.5, 1.
    \item EWC ($\alpha$): 0.001, 0.01, 0.05, 0.1, 0.5, 1, 5.
    \item LFR-CM ($\lambda$): 0.05, 0.1, 0.2, 0.4, 0.6, 0.8, 1.
    \item LFR-OM ($lr$): $ 1e-5, 2e-5, 3e-5, 4e-5, 5e-5, 1e-4$.
\end{itemize}

For the mixed-training with previous data experiments (Section 5.2), we set $\alpha$ as $0.01$ for the L2-Reg and EWC methods, $\lambda$ as $0.4$ for the LFR-CM method, and $lr$ as $1e-4$ for the LFR-OM method.

For the sequential language adaptation experiments, we set $\alpha$ as $0.5$ for the L2-Reg and EWC methods, $\lambda$ s $0.05$ for the LFR-CM method, and $lr$ as $2e-5$ for the LFR-OM method.

\end{document}